\pdfoutput=1
\documentclass[10pt,twocolumn,letterpaper]{article}

\usepackage{cvpr}              

\usepackage{graphicx}
\usepackage{amsmath}
\usepackage{amssymb}
\usepackage{booktabs}
\usepackage{multirow}
\usepackage{epsfig}
\usepackage{float}
\usepackage{caption}
\usepackage{graphicx}
\usepackage{amsmath}
\usepackage{amssymb}
\usepackage[table]{xcolor}
\usepackage{bm}
\usepackage{marvosym}
\usepackage[accsupp]{axessibility}
\usepackage{multicol}
\newlength\savewidth
\newcommand\shline{\noalign{\global\savewidth\arrayrulewidth
                            \global\arrayrulewidth 1.2pt}%
                   \hline
                   \noalign{\global\arrayrulewidth\savewidth}}
\usepackage[pagebackref,breaklinks,colorlinks]{hyperref}

\usepackage[capitalize]{cleveref}
\crefname{section}{Sec.}{Secs.}
\Crefname{section}{Section}{Sections}
\Crefname{table}{Table}{Tables}
\crefname{table}{Tab.}{Tabs.}

\def\cvprPaperID{1911} 
\def\confName{CVPR}
\def\confYear{2023}

\begin{document}

\title{UniDistill: A Universal Cross-Modality Knowledge Distillation Framework\\ for 3D Object Detection in Bird's-Eye View}

\author{Shengchao Zhou\textsuperscript{\rm 1}\thanks{Equal Contribution}\quad Weizhou Liu\textsuperscript{\rm 1}\footnotemark[1]\quad Chen Hu\textsuperscript{\rm 1}\thanks{Corresponding Author}\quad Shuchang Zhou\textsuperscript{\rm 1}\quad Chao Ma\textsuperscript{\rm 2}\vspace{5pt}\\
\textsuperscript{\rm 1} MEGVII Technology\\\textsuperscript{\rm 2} MoE Key Lab of Artificial Intelligence, AI Institute, Shanghai Jiao Tong University\\
{\tt\small \{zhoushengchao,liuweizhou,huchen,zsc\}@megvii.com},~~~{\tt\small chaoma@sjtu.edu.cn}\\\small\url{https://github.com/megvii-research/CVPR2023-UniDistill}
}
\maketitle

\begin{abstract}
In the field of 3D object detection for autonomous driving, the sensor portfolio including multi-modality and single-modality is diverse and complex. Since the multi-modal methods have system complexity while the accuracy of single-modal ones is relatively low, how to make a tradeoff between them is difficult. In this work, we propose a universal cross-modality knowledge distillation framework (UniDistill) to improve the performance of single-modality detectors. Specifically, during training, UniDistill projects the features of both the teacher and the student detector into Bird's-Eye-View (BEV), which is a friendly representation for different modalities. Then, three distillation losses are calculated to sparsely align the foreground features, helping the student learn from the teacher without introducing additional cost during inference. Taking advantage of the similar detection paradigm of different detectors in BEV, UniDistill easily supports LiDAR-to-camera, camera-to-LiDAR, fusion-to-LiDAR and fusion-to-camera distillation paths. Furthermore, the three distillation losses can filter the effect of misaligned background information and balance between objects of different sizes, improving the distillation effectiveness. Extensive experiments on nuScenes demonstrate that UniDistill effectively improves the mAP and NDS of student detectors by 2.0\%$\sim$3.2\%.
\end{abstract}
\begin{figure}[t]
\centering
\includegraphics[width=0.8\linewidth]{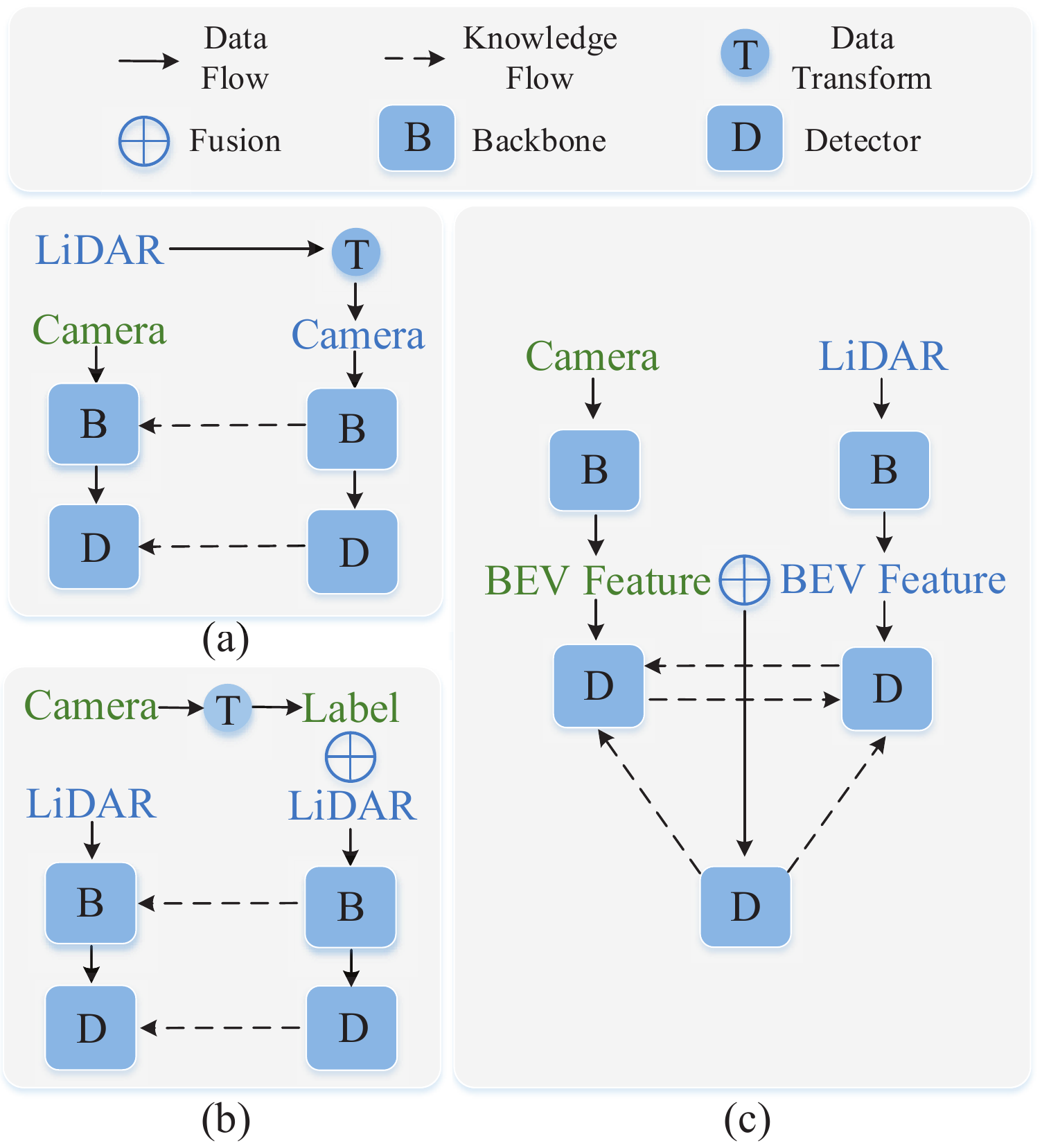}
\caption{Illustration of our proposed UniDistill. The characters in green and blue represent the data process of camera and LiDAR respectively. (a) and (b) show the procedure of two previous knowledge distillation methods, where the modalities of the teacher and the student are restricted. By contrast, our proposed UniDistill in (c) supports four distillation paths.}
\label{img:1}
\end{figure} 
\section{Introduction}
\label{sec:intro}

3D object detection plays a critical role in autonomous driving and robotic navigation. Generally, the popular 3D detectors can be categorized into (1) single-modality detectors that are based on LiDAR~\cite{shi2019pointrcnn,yang20203dssd,yin2021center,li20173d,shi2020points} or camera~\cite{huang2021bevdet,li2022bevdepth,brazil2019m3d,luo2021m3dssd} and (2) multi-modality detectors~\cite{qi2018frustum,liu2022bevfusion,vora2020pointpainting,wang2021pointaugmenting} that are based on both modalities. By fusing the complementary knowledge of two modalities, multi-modality detectors outperform their single-modality counterparts. Nevertheless, simultaneously processing the data of two modalities unavoidably introduces extra network designs and computational overhead. Worse still, the breakdown of any modality directly fails the detection, hindering the application of these detectors.

As a solution, some recent works introduced knowledge distillation to transfer complementary knowledge of other modalities to a single-modality detector. In~\cite{chong2021monodistill,ju2022paint,zheng2022boosting}, as illustrated in Figure \ref{img:1}(a) and \ref{img:1}(b), for a single-modality student detector, the authors first performed data transformation of different modalities to train a structurally identical teacher. The teacher was then leveraged to transfer knowledge by instructing the student to produce similar features and prediction results. In this way, the single-modality student obtains multi-modality knowledge and improves performance, without additional cost during inference.

Despite their effectiveness to transfer cross-modality knowledge, the application of existing methods is limited since the modalities of both the teacher and the student are restricted. In~\cite{chong2021monodistill}, the modalities of the teacher and student are fixed to be LiDAR and camera while in~\cite{zheng2022boosting,ju2022paint}, they are determined to be LiDAR-camera and LiDAR. However, the sensor portfolio in the field of 3D object detection results in a diverse and complex application of different detectors. With restricted modalities of both the teacher and student, these methods are difficult to be applied in more situations, \eg, the method in~\cite{chong2021monodistill} is not suitable to transfer knowledge from a camera based teacher to a LiDAR based student.

To solve the above problems, we propose a universal cross-modality knowledge distillation framework (UniDistill) that helps single-modality detectors improve performance. Our motivation is based on the observation that the detectors of different modalities adopt a similar detection paradigm in bird's-eye view (BEV), where after transforming the low-level features to BEV, a BEV encoder follows to further encode high-level features and a detection head produces response features to perform final prediction. 

UniDistill takes advantage of the similarity to construct the universal knowledge distillation framework. As in Figure \ref{img:1}(c), during training, UniDistill projects the features of both the teacher and the student detector into the unified BEV domain. Then for each ground truth bounding box, three distillation losses are calculated to transfer knowledge: (1) A feature distillation loss that transfers the semantic knowledge by aligning the low-level features of 9 crucial points. (2) A relation distillation loss that transfers the structural knowledge by aligning the relationship between the high-level features of 9 crucial points. (3) A response distillation loss that closes the prediction gap by aligning the response features in a Gaussian-like mask. Since the aligned features are commonly produced by different detectors, UniDistill easily supports LiDAR-to-camera, camera-to-LiDAR, fusion-to-LiDAR and fusion-to-camera distillation paths. Furthermore, the three losses sparsely align the foreground features to filter the effect of misaligned background information and balance between objects of different scales, improving the distillation effectiveness.

In summary, our contributions are three-fold:
\begin{itemize}
    \item We propose a universal cross-modality knowledge distillation framework (UniDistill) in the friendly BEV domain for single-modality 3D object detectors. With the transferred knowledge of different modalities, the performance of single-modality detectors is improved without additional cost during inference.
    \item Benefiting from the similar detection paradigm in BEV, UniDistill supports LiDAR-to-camera, camera-to-LiDAR, fusion-to-LiDAR and fusion-to-camera distillation paths. Moreover, three distillation losses are designed to sparsely align foreground features, filtering the effect of background information misalignment and balance between objects of different sizes.
    \item Extensive experiments on nuScenes demonstrate that UniDistill can effectively improve the mAP and NDS of student detectors by 2.0\%$\sim$3.2\%.
\end{itemize}
	\begin{figure*}[ht]
\centering
\includegraphics[width=0.8\linewidth,height=7.7cm]{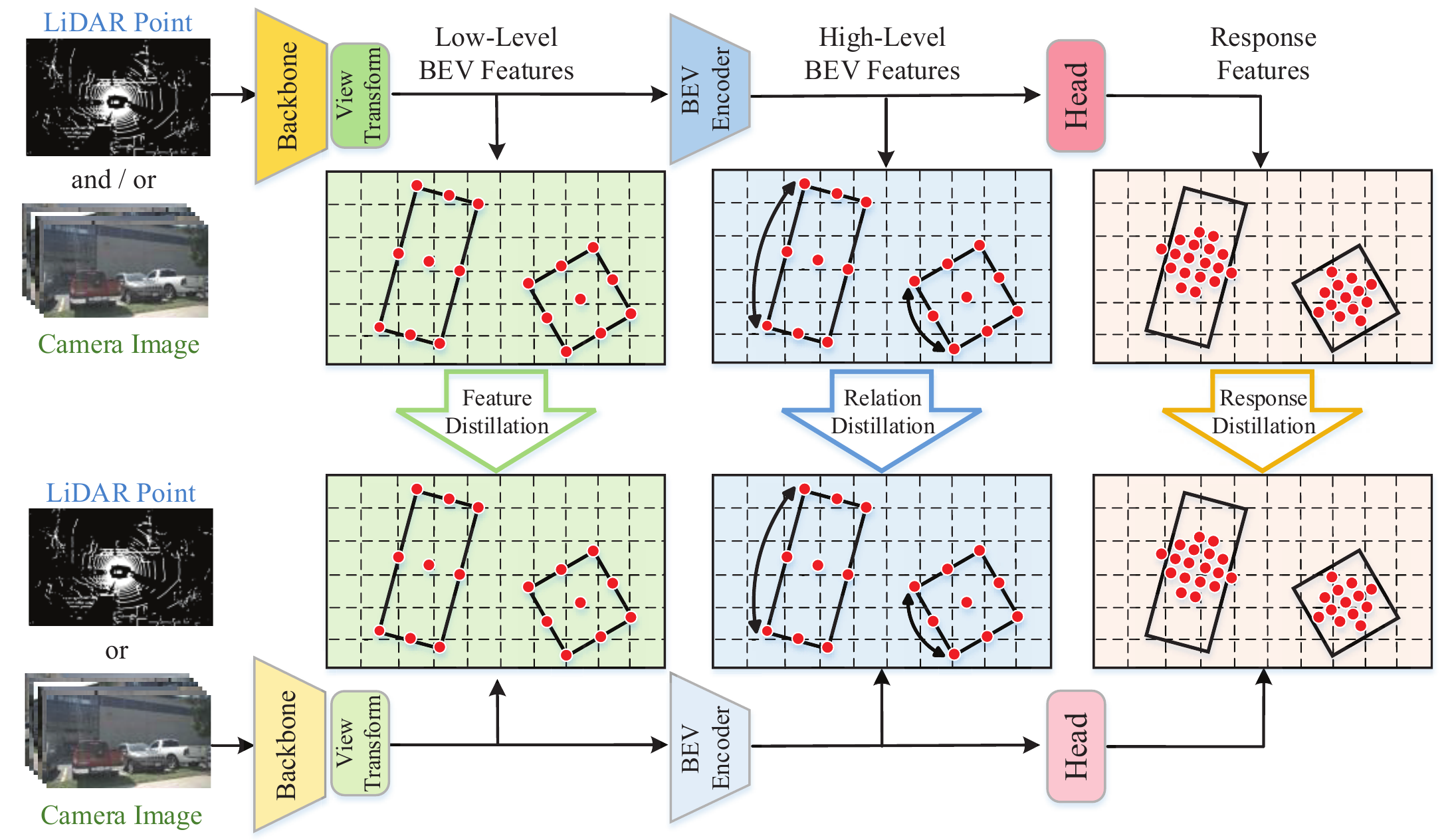}
\caption{The overview of our proposed UniDistill. The branches on the top and bottom are the teacher detector and student detector respectively. For a teacher and a student, three distillation losses to align specific foreground features are calculated after extracting the low-level features and transforming them to BEV, which are feature distillation, relation distillation and response distillation. }
\label{img:2}
\end{figure*} 
\section{Related Work}
\label{sec:relatedwork}
\subsection{3D Object Detection}
Recent mainstream 3D object detectors can be generally divided into two categories: (1) Single-modality detectors based on LiDAR~\cite{pan20213d,liu2020tanet,yin2021center,mao2021pyramid,shi2020pv} or camera~\cite{huang2021bevdet,li2022bevdepth,wang2021fcos3d,wang2022detr3d,park2021pseudo,wang2019pseudo} (2) Multi-modality detectors~\cite{qi2018frustum,liu2022bevfusion,yoo20203d,huang2020epnet} with both types of data as input. One category of LiDAR based detectors are grid-based~\cite{yan2018second,lang2019pointpillars,zhou2018voxelnet,mao2021voxel,yin2021center}, where the unstructured LiDAR points are first distributed to regular grids. As the seminal work, VoxelNet~\cite{zhou2018voxelnet} voxelizes the point clouds, performs 3D convolution, reshapes the features into BEV features and then proposes bounding boxes. PointPillars~\cite{lang2019pointpillars} substitutes the voxels with pillars to encode point clouds, avoiding time-consuming 3D convolution operations. CenterPoint~\cite{yin2021center} proposes an anchor-free detection head and obtains better performance. 

Most camera based detectors perform detection in perspective view like 2D detection. In~\cite{wang2021fcos3d}, the authors proposed FCOS3D, which is an extension of FCOS~\cite{tian2019fcos} to the field of 3D object detection. Similarly in~\cite{wang2022detr3d}, following DETR~\cite{carion2020end}, the authors proposed DETR3D to perform detection in an attention pattern. Recently, some works~\cite{huang2021bevdet,li2022bevdepth} are proposed to detect objects in BEV by applying Lift-Splat-Shoot~\cite{philion2020lift} to transform image features from perspective view to BEV and achieve satisfactory improvement.

LiDAR-camera based detectors outperform the above single-modality counterparts by fusing the complementary knowledge of LiDAR points and images. With 2D detection results, F-PointNet~\cite{qi2018frustum} obtains candidate object areas, gathers the points inside and then performs LiDAR based detection. AVOD~\cite{ku2018joint} and MV3D~\cite{chen2017multi} perform modality fusion with object proposals via ROI pooling. UVTR~\cite{li2022uvtr} unifies multi-modality representations in the voxel space for transformer-based 3D object detection. A recent state-of-the-art detector BEVFusion~\cite{liu2022bevfusion} proposes to transform both the image and the LiDAR features to BEV for modality fusion and result prediction.

Our work is inspired by that the recent detectors adopt a similar detection paradigm in BEV. After transforming the features to BEV, a similar procedure follows, where a BEV encoder further encodes high-level features and a detection head produces prediction results. The similarity of the detection paradigm makes cross-modality knowledge distillation in a universal framework possible.
\subsection{Knowledge Distillation}
Knowledge distillation is initially proposed in \cite{hinton2015distilling} for model compression and the main idea is transferring the learned knowledge from a teacher network to a student. Different works have different interpretations of the knowledge, which include the soft targets of the output layer~\cite{hinton2015distilling} and the intermediate feature map~\cite{romero2014fitnets}. Because of the effectiveness of knowledge distillation, it has been widely investigated in a variety of computer vision tasks, such as 2D object detection~\cite{dai2021general,guo2021distilling,zhang2020improve} and semantic segmentation~\cite{hou2020inter,liu2019structured}.

Recently, it is introduced into 3D object detection for knowledge transfer to single-modality detectors. In~\cite{chong2021monodistill}, the authors proposed to transfer the depth knowledge of LiDAR points to a camera based student detector by training another camera based teacher with LiDAR projected to perspective view. In~\cite{zheng2022boosting,ju2022paint}, a PointPainting~\cite{vora2020pointpainting} teacher is leveraged to instruct a CenterPoint student to produce similar features and responses. 

Although these methods are effective for knowledge transfer, the modalities of the teacher and the student are restricted. However, the diverse and complex application of different detectors will restrict their application. Instead, our proposed UniDistill projects the features of detectors to BEV and supports LiDAR-to-camera, camera-to-LiDAR, fusion-to-LiDAR and fusion-to-camera distillation paths.

We note that a recent work BEVDistill also performs cross-modality distillation in BEV. However, it also imposes restrictions on the modalities of the teacher and the student, resulting in a limited application. Moreover, their main distillation losses are designed for transformer based detectors. Differently, our UniDistill aims to improve the performance of CNN based detectors.
	\section{Methodology}

In this section, we describe our proposed UniDistill in detail. In Section \ref{sec:3.1}, we briefly introduce the similar detection paradigm of different detectors in BEV, where some low-level BEV features are obtained via view-transform and further encoded to be high-level features and response features.  In Section \ref{sec:3.2}, we depict the proposed UniDistill framework for cross-modality knowledge distillation. For a teacher and a student detector, after obtaining their features in BEV, three distillation losses aligning foreground features are calculated to perform knowledge transfer.

\subsection{Preliminary}
\label{sec:3.1}
The recent detectors adopt a similar detection paradigm in BEV, where the major difference is the approach to get the low-level BEV features. For the camera based detectors proposed in~\cite{huang2021bevdet,li2022bevdepth}, after pixel-wise depth estimation, the image features are projected from the perspective view into BEV via the view-transform operation proposed in~\cite{philion2020lift} to form low-level BEV features $\boldmath{F}_{\text{cam}}^{\text{low}}$. 

With respect to LiDAR based detectors presented in \cite{yin2021center,zhou2018voxelnet,lang2019pointpillars}, the unstructured point clouds are first distributed to regular voxels or pillars. The features of voxels or pillars are then extracted and reshaped into low-level BEV features $\boldmath{F}_{\text{ldr}}^{\text{low}}$ by concatenating the voxel features in the same column. 

In \cite{liu2022bevfusion}, the authors proposed a method to construct a LiDAR-camera based detector simply by fusing the low-level features $\boldmath{F}_{\text{ldr}}^{\text{low}}$ and $\boldmath{F}_{\text{cam}}^{\text{low}}$ of a LiDAR and camera based detector. It first concatenates $\boldmath{F}_{\text{ldr}}^{\text{low}}$ and $\boldmath{F}_{\text{cam}}^{\text{low}}$ and then processes the result with a fully convolutional network to produce the fused features $\boldmath{F}_{\text{fuse}}^{\text{low}}$.

The following steps are the same for different detectors. For a detector of any modality, a BEV encoder (BEVEnc) first takes its low-level BEV features $\boldmath{F}_{\text{mod}}^{\text{low}}$ as input to further encode the high-level features $\boldmath{F}_{\text{mod}}^{\text{high}}$:
\begin{equation}
\boldmath{F}_{\text{mod}}^{\text{high}} = \text{BEVEnc}(\boldmath{F}_{\text{mod}}^{\text{low}}),
\end{equation}
where mod is the detector modality and is in $\{\text{ldr,cam,fuse}\}$. Then a detection head (DetHead) produces the classification and regression heatmaps $\boldmath{F}_{\text{mod}}^{\text{cls}}$ and $\boldmath{F}_{\text{mod}}^{\text{reg}}$, based on which the final predictions are generated:
\begin{equation}
\boldmath{F}_{\text{mod}}^{\text{cls}}, \boldmath{F}_{\text{mod}}^{\text{reg}} = \text{DetHead}(\boldmath{F}_{\text{mod}}^{\text{high}}).
\end{equation}

Therefore, regardless of the modality, these detectors will consistently produce $\boldmath{F}^{\text{low}}$, $\boldmath{F}^{\text{high}}$, $\boldmath{F}^{\text{cls}}$ and $\boldmath{F}^{\text{reg}}$ during the procedure of detection.

\subsection{UniDistill}
\label{sec:3.2}
Taking advantage of the above similar detection paradigm in BEV, we propose a universal cross-modality knowledge distillation framework (UniDistill), which easily supports LiDAR-to-camera, camera-to-LiDAR, fusion-to-LiDAR and fusion-to-camera distillation paths. The overview of UniDistill is illustrated in Figure \ref{img:2}. During training, after transforming the low-level features of the teacher and the student of different modalities to BEV, three distillation losses are then calculated. The losses are finally combined with the original detection loss to train the student. In this way, the student can mimic the teacher to learn cross-modal knowledge and thus gain better detection results, without introducing additional cost during inference.

Denoting the modalities of the teacher and the student as MT and MS respectively, the three distillation losses can be interpreted as follows: (1) The first loss “feature distillation” transfers the semantic knowledge in the low-level BEV features $\boldmath{F}_{\text{MT}}^{\text{low}}$ to $\boldmath{F}_{\text{MS}}^{\text{low}}$ by point-wisely aligning the features of 9 crucial points of each ground truth bounding box. (2) The second loss “relation distillation” transfers the structural knowledge in the high-level BEV features $\boldmath{F}_{\text{MT}}^{\text{high}}$ to $\boldmath{F}_{\text{MS}}^{\text{high}}$ by group-wisely aligning the relationship between 9 crucial points of each ground truth bounding box. (3) The third loss “response distillation” closes the prediction gap by aligning the heatmaps $(\boldmath{F}_{\text{MS}}^{\text{cls}},\boldmath{F}_{\text{MS}}^{\text{reg}})$ with $(\boldmath{F}_{\text{MT}}^{\text{cls}},\boldmath{F}_{\text{MT}}^{\text{reg}})$ in a Gaussian-like mask for each ground truth bounding box.
\subsubsection{Feature Distillation}
\quad Since the low-level BEV features provide semantic knowledge for further process, we propose feature distillation to align $\boldmath{F}_{\text{MS}}^{\text{low}}$ with $\boldmath{F}_{\text{MT}}^{\text{low}}$. One intuitive method is completely aligning $\boldmath{F}_{\text{MS}}^{\text{low}}$ with $\boldmath{F}_{\text{MT}}^{\text{low}}$, however, the background information misalignment between different modalities will decrease the  effectiveness. Worse still, because different objects occupy areas of different sizes, it will focus more on aligning the features of large objects than small objects. 

To mitigate the above effects, in feature distillation, we only align the features of foreground objects and equally select 9 crucial points for each of them for alignment. Specifically, in BEV, the bounding box of one foreground object can be regarded as a 2D rotated box and described by the coordinates of its corners $\{(x_i,y_i)\}_{i=1}^4$. Based on the four corners, the coordinates of midpoints of 4 edges and the center of the box can be calculated. We collect these 9 points together as $\{(x_i,y_i)\}_{i=1}^9$ and regard them as crucial points for the bounding box. For each point $\bm{p}_i$=$(x_i,y_i)$, we calculate the difference between its features on $\boldmath{F}_{\text{MT}}^{\text{low}}$ and $\boldmath{F}_{\text{MS}}^{\text{low}}$ to form the feature distillation loss $\mathcal{L}_{\text{Fea}}$:
\begin{equation}
\mathcal{L}_{\text{Fea}} = (\sum_{i=1}^9|\bm{F}_{\text{MT}}^{\text{low}}(x_i, y_i)-\bm{F}_{\text{MS}}^{\text{low}}(x_i, y_i)|)/9.
\end{equation}

In addition, we note that when the performance of the teacher detector is worse than the student, \eg, the modalities of the teacher and the student are camera and LiDAR, using the vanilla feature distillation can even degrade the final performance. Inspired by the feature adaptation operation in ~\cite{romero2014fitnets,chen2017learning}, in this situation, we also introduce an adaptive layer $\text{Adapt}_{1}$, which is a one-layer convolutional network, after $\boldmath{F}_{\text{MS}}^{\text{low}}$ to produce new features $\hat{\bm{F}}_{\text{MS}}^{\text{low}}$ and calculate feature distillation instead with $\hat{\bm{F}}_{\text{MS}}^{\text{low}}$ and $\boldmath{F}_{\text{MT}}^{\text{low}}$. During inference, the adaptive layer is removed and the original low-level feature $\boldmath{F}_{\text{MS}}^{\text{low}}$ is further processed to generate predictions. Therefore, there is no modification to the structure of the student detector.
\subsubsection{Relation Distillation}
\quad The high-level features can provide knowledge about the structure of the scene, which is important for the final prediction. To transfer the structural knowledge from the teacher to the student, we propose relation distillation to align the relationship between the different parts of an object in $\bm{F}_{\text{MS}}^{\text{high}}$ and $\bm{F}_{\text{MT}}^{\text{high}}$. Specifically, for one ground truth bounding box, we also consider its 9 crucial points $\{(x_i,y_i)\}_{i=1}^9$ the same as those in feature distillation. We first gather their features on $\bm{F}_{\text{MS}}^{\text{high}}$ and calculate the relationship between them to form a relation matrix $\text{RelMat}^{\text{MS}}$ with the size of 9$\times$9:
\begin{equation}
\text{RelMat}^{\text{MS}}_{i,j} = \Phi(\bm{F}_{\text{MS}}^{\text{high}}(x_{i},y_{i}),\bm{F}_{\text{MS}}^{\text{high}}(x_j,y_{j})), 
\end{equation}
where 1$\leqslant$$i,j$$\leqslant$9 and $\Phi$ represents the cosine similarity function. With the same operation, another relation matrix $\text{RelMat}^{\text{MT}}$ can be calculated based on $\bm{F}_{\text{MT}}^{\text{high}}$ and then the relation distillation loss $\mathcal{L}_{\text{Rel}}$ is calculated to completely align $\text{RelMat}^{\text{MS}}$ with $\text{RelMat}^{\text{MT}}$:
\begin{equation}
\mathcal{L}_{\text{Rel}} = (\sum_{1\leqslant i,j \leqslant 9}|\text{RelMat}^{\text{MT}}_{i,j}-\text{RelMat}^{\text{MS}}_{i,j}|)/81.
\end{equation}

Furthermore, the same as that in feature distillation, when the teacher performs worse than student, we introduce another one-layer convolutional network as an adaptive layer $\text{Adapt}_2$ after $\bm{F}_{\text{MS}}^{\text{high}}$ to produce new features $\hat{\bm{F}}_{\text{MS}}^{\text{high}}$ and calculate relation distillation with $\hat{\bm{F}}_{\text{MS}}^{\text{high}}$ and $\bm{F}_{\text{MT}}^{\text{high}}$.

\subsubsection{Response Distillation}
\quad To make the final prediction of the student similar to the teacher, we further propose response distillation. Based on the high-level BEV features, the detection head will produce a classification heatmap $\bm{F}^{\text{cls}}\in R^{H\times W\times C}$ and a regression heatmap $\bm{F}^{\text{reg}}\in R^{H\times W\times T}$, where $H$ and $W$ are the height and width of the heatmap, $C$ is the number of classes to be predicted and $T$ is the number of regression targets. We further gather the max value of each position in $\bm{F}^{\text{cls}}$ to form a new heatmap $\bm{F}_{\text{max}}^{\text{cls}}$:
\begin{equation}
\bm{F}_{\text{max}}^{\text{cls}}(i,j)=\max_{1\leqslant k\leqslant C}\bm{F}^{\text{cls}}(i,j,k),
\end{equation}
where 1$\leqslant$$i$$\leqslant$$H$ and 1$\leqslant$$j$$\leqslant$$W$. Then $\bm{F}_{\text{max}}^{\text{cls}}$ is concatenated with $\bm{F}^{\text{reg}}$ to be the response features $\bm{F}^{\text{resp}}$. In this way, we obtain the response features of the teacher and the student, which are $\bm{F}_{\text{MT}}^{\text{resp}}$ and $\bm{F}_{\text{MS}}^{\text{resp}}$ respectively,  and calculate response distillation loss to align $\bm{F}_{\text{MS}}^{\text{resp}}$ with $\bm{F}_{\text{MT}}^{\text{resp}}$. To mitigate the effect of background information misalignment, we only align the part of response features near the foreground objects. Instead of selecting 9 crucial points, we find that the quality of the response values in $\bm{F}_{\text{MT}}^{\text{resp}}$ near the center of a ground truth bounding box is good enough to guide $\bm{F}_{\text{MS}}^{\text{resp}}$. Therefore, for one ground truth bounding box, we generate a Gaussian-like mask with the same method in \cite{chong2021monodistill}, gather the response values inside the mask and calculate response distillation loss $\mathcal{L}_{\text{Resp}}$ to align them:
\begin{equation}
\mathcal{L}_{\text{Resp}} = \sum_{1\leqslant i\leqslant \text{H},1\leqslant j\leqslant \text{W}}|\bm{F}_{\text{MT}}^{\text{resp}}(i,j)-\bm{F}_{\text{MS}}^{\text{resp}}(i,j)|\times \text{Mask}(i,j),
\end{equation}
where $\text{Mask}$ is the calculated Gaussian-like mask and only the area near the ground truth bounding box is non-zero.
\begin{table*}[ht]\small
\centering
\caption{Performance analysis of UniDistill in four distillation paths on the nuScenes test dataset. “*” indicates re-implementation on our new student detector. “L” and “C” represent the LiDAR and camera.}
\resizebox{0.92\width}{!}{
\begin{tabular}{c|cc|ccccccc}
\shline
Method      & Modality & \begin{tabular}[c]{@{}c@{}}Teacher\\ Modality\end{tabular} & mAP~$\uparrow$ & NDS~$\uparrow$                           & mATE~$\downarrow$                                & mASE~$\downarrow$                                & mAOE~$\downarrow$                                & mAVE~$\downarrow$                                  & mAAE~$\downarrow$                                \\ \shline
CVCNet\cite{chen2020every}      & L        & -                                                          & 55.8                & 64.2                & 30.0                & 24.8                & 43.1                & 26.9                  & 11.9                \\
Guided 3DOD\cite{fazlali2022versatile} & L        & -                                                          & 60.9                & 67.3                & 28.8                & 24.5                & 40.0                & 25.3                  & 12.8                \\
AFDetV2\cite{hu2022afdetv2}    & L        & -                                                          & 62.4                & 68.5                & 25.7                & 23.4                & 34.1                & 29.9                  & 13.7                \\
S2M2-SSD*\cite{zheng2022boosting}    & L        & L+C                                                        & 63.6                & 69.6                & 25.4                & 23.9                & 34.3                & 25.7                  & 12.8                \\ \shline
UniDistill  & L+C      & -                                                          & 65.4                & 70.6                & 25.1                & 23.8                & 32.5                & 25.6                 & 12.8                \\ \shline
UniDistill  & L        & -                                                          & 61.4                & 67.8                & 26.8                & 25.1                & 33.6                & 27.8                  & 14.8                \\
\rowcolor{black!5}UniDistill  & L        & C                                                          & \textbf{63.4(+2.0)} & \textbf{69.8(+2.0)} & \textbf{24.9(-1.9)} & \textbf{23.7(-1.4)} & \textbf{32.1(-1.5)} & \textbf{24.7(-3.1)}   & \textbf{13.1(-1.7)} \\
\rowcolor{black!10}UniDistill  & L        & L+C                                                        & \textbf{63.9(+2.5)} & \textbf{70.1(+2.3)} & \textbf{25.0(-1.8)} & \textbf{23.8(-1.3)} & \textbf{32.8(-0.8)} & \textbf{24.5(-3.3)}   & \textbf{12.7(-2.1)} \\ \shline
UniDistill  & C        & -                                                          & 26.4                & 36.1                & 69.7                & 26.6                & 55.8                & 117.3                 & 17.8                \\
\rowcolor{black!5}UniDistill  & C        & L                                                          & \textbf{28.9(+2.5)} & \textbf{38.4(+2.3)} & \textbf{65.9(-3.8)} & \textbf{25.9(-0.7)} & \textbf{51.4(-4.4)} & \textbf{106.4(-10.9)} & \textbf{17.0(-0.8)} \\
\rowcolor{black!10}UniDistill  & C        & L+C                                                        & \textbf{29.6(+3.2)} & \textbf{39.3(+3.2)} & \textbf{63.7(-6.0)} & \textbf{25.7(-0.9)} & \textbf{49.2(-6.6)} & \textbf{108.4(-8.9)}  & \textbf{16.7(-1.1)} \\ \shline
\end{tabular}
}
\label{tab:1}
\end{table*}

\subsubsection{Total Objectives}
\quad After being calculated for each ground truth bounding box, every distillation loss is then averaged over all bounding boxes. Finally, we combine the detection loss $\mathcal{L}_{\text{Det}}$ of the student with the distillation losses as the total loss $\mathcal{L}_{\text{Total}}$:
\begin{equation}
\mathcal{L}_{\text{Total}} = \mathcal{L}_{\text{Det}}+\lambda_1\cdot\mathcal{L}_{\text{Fea}}+\lambda_2\cdot\mathcal{L}_{\text{Rel}}+\lambda_3\cdot\mathcal{L}_{\text{Resp}},
\end{equation}
where $\lambda_1$, $\lambda_2$ and $\lambda_3$ are hyperparameters used to balance the scale of different losses. The student is then optimized by $\mathcal{L}_{\text{Total}}$, achieving better performance.

	\section{Experiments}
\subsection{Experimental Setup}
\subsubsection{Dataset}
\quad We choose the popular large-scale dataset nuScenes for evaluation in our experiments. This dataset consists of 1,000 driving sequences (700/150/150 for train/val/test) with images from 6 cameras, points from 5 Radars and 1 LiDAR. Each sequence is 20 seconds long and every 10 frames are fully annotated with 3D object bounding boxes. For each frame, there are 6 RGB camera images and 20 LiDAR scanning frames. Following BEVFusion~\cite{liu2022bevfusion}, we set the area inside [-54.0\text{m}, 54.0\text{m}]$\times$[-54.0\text{m}, 54.0\text{m}]$\times$[-5\text{m},  3\text{m}] as our region of interest.
\subsubsection{Evaluation Metrics}
\quad We use the official evaluation metrics of nuScenes, which are the mean Average Precision(mAP), mean Average Translation Error(mATE), mean Average Scale Error(mASE), mean Average Orientation Error(mAOE), mean Average Velocity Error(mAVE), mean Average Attribute Error(mAAE) and the nuScenes detection score(NDS). The mAP measures the recall and precision of predicted bounding boxes. NDS is the composition of other metrics to comprehensively judge the detection capacity. The remaining metrics calculate the positive results of prediction on the corresponding aspects. The data unit of results is “\%”.

\subsubsection{Models and Training}
\quad We adopt BEVDet~\cite{huang2021bevdet} with ResNet-50 as our camera based detector and CenterPoint\cite{yin2021center} as our LiDAR based detector. The LiDAR-camera detector is constructed by fusing the low-level BEV features of BEVDet and CenterPoint, following the method proposed in BEVFusion~\cite{liu2022bevfusion}. The size of voxel to produce BEV features is 0.075\text{m}$\times$0.075\text{m}$\times$0.2\text{m}.

The detectors are trained by an AdamW optimizer with the learning rate to be 1e-4 for LiDAR and LiDAR-camera based detectors and 2e-4 for the camera based detector. The training batch size is 20 and the training epoch is 20 for all detectors. More details are supplemented in the appendix.

UniDistill is evaluated to see whether it can transfer knowledge in four distillation paths: (1) From the LiDAR-camera based teacher detector to the LiDAR based student. (2) From the LiDAR-camera based teacher detector to the camera based student. (3) From the camera based teacher detector to the LiDAR based student. (4) From the LiDAR based teacher detector to the camera based student. The hyperparameters $\lambda_1$,$\lambda_2$ and $\lambda_3$ in each path are: (1) 10, 1, 10. (2) 10, 5, 10. (3) 10, 5, 1. (4) 100, 40, 10. The adaptive layers for feature distillation and relation distillation are introduced when evaluating in path (3).


\subsection{Comparison with the State-of-the-Arts}
We first evaluate the performance of UniDistill on the test dataset of nuScenes and Table \ref{tab:1} reports the results. It is revealed that in all of the four distillation paths, UniDistill helps transfer knowledge from the teacher detector to student and improve its performance. Besides, with the LiDAR-camera based detector as the teacher, the LiDAR based student obtains better performance than other state-of-the-art LiDAR based detectors, proving the effectiveness of UniDistill. We also compare UniDistill with S2M2-SSD~\cite{zheng2022boosting}, which performs cross-modality knowledge distillation from a PointPainting~\cite{vora2020pointpainting} teacher detector also to a CenterPoint student detector. The result shows that UniDistill helps the student obtain better performance.
\begin{table*}[ht]\small
\centering
\caption{Ablation study of three proposed distillation losses on the nuScenes validation dataset. “T” and “S” represent the teacher detector and student detector respectively and the mAP/NDS of the student is reported.}
\begin{tabular}{c|ccc|cccc}
\shline
\multirow{2}{*}{Setting} & \multicolumn{3}{c|}{Loss} & \multicolumn{4}{c}{Modality}                                                                                                                                                                                                                                                                                                  \\ \cline{2-8} 
                     & $\mathcal{L}_{\text{Fea}}$       & $\mathcal{L}_{\text{Rel}}$      & $\mathcal{L}_{\text{Resp}}$      & \multicolumn{1}{c|}{\begin{tabular}[c]{@{}c@{}}T:LiDAR+Camera\\ S:LiDAR\end{tabular}} & \multicolumn{1}{c|}{\begin{tabular}[c]{@{}c@{}}T:LiDAR+Camera\\ S:Camera\end{tabular}} & \multicolumn{1}{c|}{\begin{tabular}[c]{@{}c@{}}T:Camera\\ S:LiDAR\end{tabular}} & \begin{tabular}[c]{@{}c@{}}T:LiDAR\\ S:Camera\end{tabular} \\ \shline
1                    &         &        &        & \multicolumn{1}{c|}{53.5/63.9}                                                        & \multicolumn{1}{c|}{20.3/33.1}                                                         & \multicolumn{1}{c|}{53.5/63.9}                                                  & 20.3/33.1                                                  \\
2                    & \checkmark       &        &        & \multicolumn{1}{c|}{56.1/65.5}                                                        & \multicolumn{1}{c|}{21.6/34.5}                                                         & \multicolumn{1}{c|}{54.3/64.6}                                                  & 21.1/34.3                                                  \\
3                    &         & \checkmark      &        & \multicolumn{1}{c|}{54.1/64.0}                                                        & \multicolumn{1}{c|}{22.3/35.7}                                                         & \multicolumn{1}{c|}{55.2/65.3}                                                  & 21.7/35.0                                                  \\
4                    &         &        & \checkmark      & \multicolumn{1}{c|}{58.7/66.7}                                                        & \multicolumn{1}{c|}{25.7/37.1}                                                         & \multicolumn{1}{c|}{55.7/65.6}                                                  & 24.9/36.3                                                  \\
5                    & \checkmark       & \checkmark      &        & \multicolumn{1}{c|}{-}                                                                & \multicolumn{1}{c|}{-}                                                                 & \multicolumn{1}{c|}{-}                                                          & 23.5/35.4                                                  \\
6                    &         & \checkmark      & \checkmark      & \multicolumn{1}{c|}{-}                                                                & \multicolumn{1}{c|}{-}                                                                 & \multicolumn{1}{c|}{-}                                                          & 25.3/36.7                                                  \\
7                    & \checkmark       &        & \checkmark      & \multicolumn{1}{c|}{-}                                                                & \multicolumn{1}{c|}{-}                                                                 & \multicolumn{1}{c|}{-}                                                          & 25.3/37.0                                                  \\
8                    & \checkmark       & \checkmark      & \checkmark      & \multicolumn{1}{c|}{\textbf{59.7/67.5}}                                               & \multicolumn{1}{c|}{\textbf{26.5/37.8}}                                                & \multicolumn{1}{c|}{\textbf{57.0/66.3}}                                         & \textbf{26.0/37.3}                                         \\ \shline
\end{tabular}
\label{tab:2}
\end{table*}

\subsection{Ablation Studies}
In this section, some experiments are conducted on the validation dataset to show the effect of each distillation loss and the rationality of specific designs. For efficiency, we turn off the auto-scaling between classification and regression loss in $\mathcal{L}_{\text{Det}}$. We report the overall results in Table \ref{tab:2} to show the effect of each loss.
\subsubsection{Effect of Feature Distillation}
\quad As in the second setting of Table \ref{tab:2}, using the feature distillation improves the NDS and mAP of the student in four paths. Moreover, the fifth, seventh and eighth settings prove that it is complementary to other distillation losses. 

Another experiment in path (4) is conducted to show the rationality of feature distillation to align the features of 9 crucial points. We compare the original feature distillation with two modified ones which align the low-level BEV features (1) completely or (2) inside a Gaussian-like mask like the response distillation. The results in Table \ref{tab:3} show that feature distillation performs better when selecting 9 crucial points for alignment. Moreover, in this situation, the AP of small objects (pedestrians and motors) improves a lot while the influence on large objects(cars and trucks) is minor.
\begin{table}[t]\small
\centering
\caption{Ablation study in path (4) to show that feature distillation performs better when selecting crucial points for alignment.}
\begin{tabular}{c|ccccc|c}
\shline
\multirow{2}{*}{Method} & \multicolumn{5}{c|}{AP}            & \multirow{2}{*}{NDS} \\ \cline{2-6}
                        & car  & truck & ped  & motor & mean &                      \\ \shline
Baseline                & 38.5 & 20.1  & 9.4  & 18.5  & 20.3 & 33.1                 \\
Complete                    & 38.0 & 13.1  & 14.2 & 22.0  & 20.3 & 32.6                 \\
Gaussian                & \textbf{45.3} & \textbf{21.4}  & 10.3 & 16.8  & 20.6 & 32.8                 \\ \shline
Crucial                 & 44.0 & 14.9  & \textbf{21.9} & \textbf{22.4}  & \textbf{21.1} & \textbf{34.3}                 \\ \shline
\end{tabular}
\label{tab:3}
\end{table}

\begin{table}[t]\small
\centering
\caption{Ablation study in path (4) to show that feature distillation performs better when aligning the low-level BEV features.}
\begin{tabular}{c|ccccc}
\shline
Method     & mAP  & mASE & mAOE & mAAE & NDS  \\ \shline
Baseline   & 20.3 & 27.9 & 46.6 & 21.9 & 33.1 \\
High-Level & 20.6 & 28.1 & 46.9 & 23.2 & 32.3 \\ \shline
Low-Level  & \textbf{21.1} & \textbf{27.8} & \textbf{46.3} & \textbf{21.9} & \textbf{34.3} \\ \shline
\end{tabular}
\label{tab:4}
\end{table}

We further compare the original feature distillation with a modified one which aligns the high-level BEV features of 9 crucial points in path (4). The results illustrated in Table \ref{tab:4} reveal that calculating feature distillation with the low-level BEV features obtains better performance.

\subsubsection{Effect of Relation Distillation}
\quad The comparison between the third setting and first setting of Table \ref{tab:2} shows that using the relation distillation improves the NDS and mAP of the student in all paths. And the results in the fifth, sixth and eighth settings further prove that it is complementary to other distillation losses. 

We conduct another experiment in path (4) to show the rationality of selecting 9 crucial points for relation distillation calculation. We compare the original relation distillation with two modified ones which align the relationship between (1) all of the high-level BEV features or (2) the features inside a Gaussian-like mask like the response distillation. The results illustrated in Table \ref{tab:5} show that the relation distillation obtains the best performance when calculating the relationship between 9 crucial points for alignment.

We further compare the original relation distillation with the modified one, which aligns the relationship between the low-level BEV features of 9 crucial points. The results in path (4) are illustrated in Table \ref{tab:6} and reveal that calculating relation distillation with the high-level BEV features obtains better performance.
\begin{figure*}[t]
\centering
\setlength{\belowcaptionskip}{-5pt}
\includegraphics[width=0.85\linewidth]{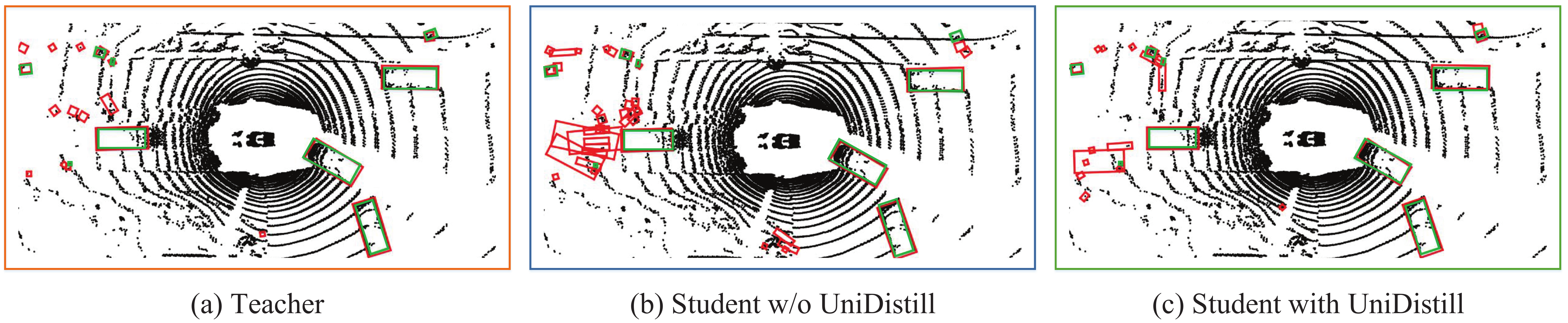}
\caption{Illustration of detection results. The boxes in red and green are the predicted and ground truth bounding boxes respectively. (a), (b) and (c) show the results of the LiDAR-camera based teacher detector, the LiDAR based student detector without UniDistill and the LiDAR based student detector with UniDistill respectively. The results show that UniDistill helps the student detector generate more accurate predictions and fewer false positive results.}
\label{img:3}
\end{figure*} 
\subsubsection{Effect of Response Distillation}
\quad As the fourth setting of Table \ref{tab:2} shows, using the response distillation improves the NDS and mAP of the student in all paths. Furthermore, the sixth, seventh and eighth settings prove that it is complementary to other distillation losses. 

Another experiment in path (4) is conducted to prove the rationality of aligning the response features inside the Gaussian-like mask for response distillation. The original response distillation is compared with two modified ones which align the response features (1) completely or (2) of 9 crucial points like those in feature distillation. The results illustrated in Table \ref{tab:7} show that the response distillation obtains better performance when selecting the response features inside the Gaussian-like mask for alignment.

To form the response features, we first gather the max value of each position in the classification heatmap to obtain a new heatmap. In path (4), we further compare the original response distillation with a modified one based on the response features formed by concatenating the original classification and regression heatmaps. The results in Table \ref{tab:8} show that gathering the max value to form the response features helps response distillation perform better.
\begin{table}[t]\small
\centering
\caption{Ablation study in path (4) to show that relation distillation performs better when selecting crucial points for alignment.}
\begin{tabular}{c|ccccc|c}
\shline
\multirow{2}{*}{Method} & \multicolumn{5}{c|}{AP}            & \multirow{2}{*}{NDS} \\ \cline{2-6}
                        & car  & truck & ped  & motor & mean &                      \\ \shline
Baseline                & 38.5 & \textbf{20.1}  & 9.4  & 18.5  & 20.3 & 33.1                 \\
Complete                     & 44.0 & 18.3  & 15.9 & 17.9  & 20.4 & 33.4                 \\
Gaussian                & \textbf{44.9} & 18.3  & 15.3 & 18.1  & 20.8 & 34.0                 \\ \shline
Crucial                 & 44.1 & 18.3  & \textbf{19.2} & \textbf{18.8}  & \textbf{21.7} & \textbf{35.0}                 \\ \shline
\end{tabular}
\label{tab:5}
\end{table}
\begin{table}[t]\small
\centering
\caption{Ablation study in path (4) to show that relation distillation performs better when aligning the high-level BEV features.}
\begin{tabular}{c|ccccc}
\shline
Method     & mAP  & mASE & mAOE & mAAE & NDS  \\ \shline
Baseline   & 20.3 & 27.9 & 46.6 & 21.9 & 33.1 \\
Low-Level  & 19.7 & \textbf{27.7} & 54.4 & 24.3 & 32.3 \\ \shline
High-Level & \textbf{21.7} & 28.1 & \textbf{46.2} & \textbf{21.2} & \textbf{35.0} \\ \shline
\end{tabular}
\label{tab:6}
\end{table}
\subsubsection{Effect of Adaptive Layers}
\begin{table}[t]\small
\centering
\caption{Ablation study in path (4) to show that response distillation performs better when aligning the features in Gaussian mask.}
\begin{tabular}{c|ccccc|c}
\shline
\multirow{2}{*}{Method} & \multicolumn{5}{c|}{AP}            & \multirow{2}{*}{NDS} \\ \cline{2-6}
                        & car  & truck & ped  & motor & mean &                      \\ \shline
Baseline                & 38.5 & 20.1  & 9.4  & 18.5  & 20.3 & 33.1                 \\
Complete                     & 45.3 & 18.8  & 13.6 & 22.1  & 23.4 & 34.9                 \\
Crucial                 & 46.7 & 17.4  & 14.6 & 22.5  & 23.4 & 35.3                 \\ \shline
Gaussian                & \textbf{47.8} & \textbf{21.1}  & \textbf{26.5} & \textbf{23.4}  & \textbf{24.9} & \textbf{36.3}                 \\ \shline
\end{tabular}
\label{tab:7}
\end{table}
\begin{table}[t]\small
\centering
\caption{Ablation study in path (4) to show that gathering the max value of each position in the classification heatmap helps response distillation perform better.}
\begin{tabular}{c|ccccc}
\shline
Method   & mAP  & mASE & mAOE & mAAE & NDS  \\ \shline
Baseline & 20.3 & 27.9 & 46.6 & \textbf{21.9} & 33.1 \\
W/O Max   & 24.3 & 27.5 & 46.7 & 22.5 & 35.4 \\ \shline
Max      & \textbf{24.9} & \textbf{27.1} & \textbf{42.6} & 24.0 & \textbf{36.3} \\ \shline
\end{tabular}
\label{tab:8}
\end{table}
\begin{table}[t]\small
\centering
\caption{Ablation study in path (3) to show that the adaptive layers help feature distillation and relation distillation perform better.}
\resizebox{0.92\width}{!}{
\begin{tabular}{c|ccc|ccc}
\shline
\multirow{2}{*}{Method} & \multicolumn{3}{c|}{$\mathcal{L}_{\text{Fea}}$} & \multicolumn{3}{c}{$\mathcal{L}_{\text{Rel}}$} \\ \cline{2-7} 
                        & mAP    & mAVE  & NDS   & mAP   & mAVE  & NDS   \\ \shline
Baseline                & 53.5   & 22.5  & 63.9  & 53.5  & 22.5  & 63.9  \\
W/O Adapt                & 53.3   & 25.3  & 63.5  & 53.1  & 24.3  & 63.2  \\ \shline
With Adapt              & \textbf{54.3}   & \textbf{21.3}  & \textbf{64.6}  & \textbf{55.2}  & \textbf{21.4}  & \textbf{65.3}  \\ \shline
\end{tabular}
}
\label{tab:9}
\end{table}
\quad When evaluating in path (3), we introduce two adaptive layers $\text{Adapt}_1$ and $\text{Adapt}_2$ after the low- and high-level BEV features to avoid performance degradation. Here we conduct experiments to show the indispensability of these adaptive layers by removing them and re-evaluating the performance. The results in Table \ref{tab:9} show that without the adaptive layers, the feature distillation and relation distillation even worsen the performance of the student detector. Therefore, in this situation, we keep the two adaptive layers for help.

\begin{figure}[t]
\centering
\setlength{\belowcaptionskip}{-10pt}
\includegraphics[width=0.95\linewidth]{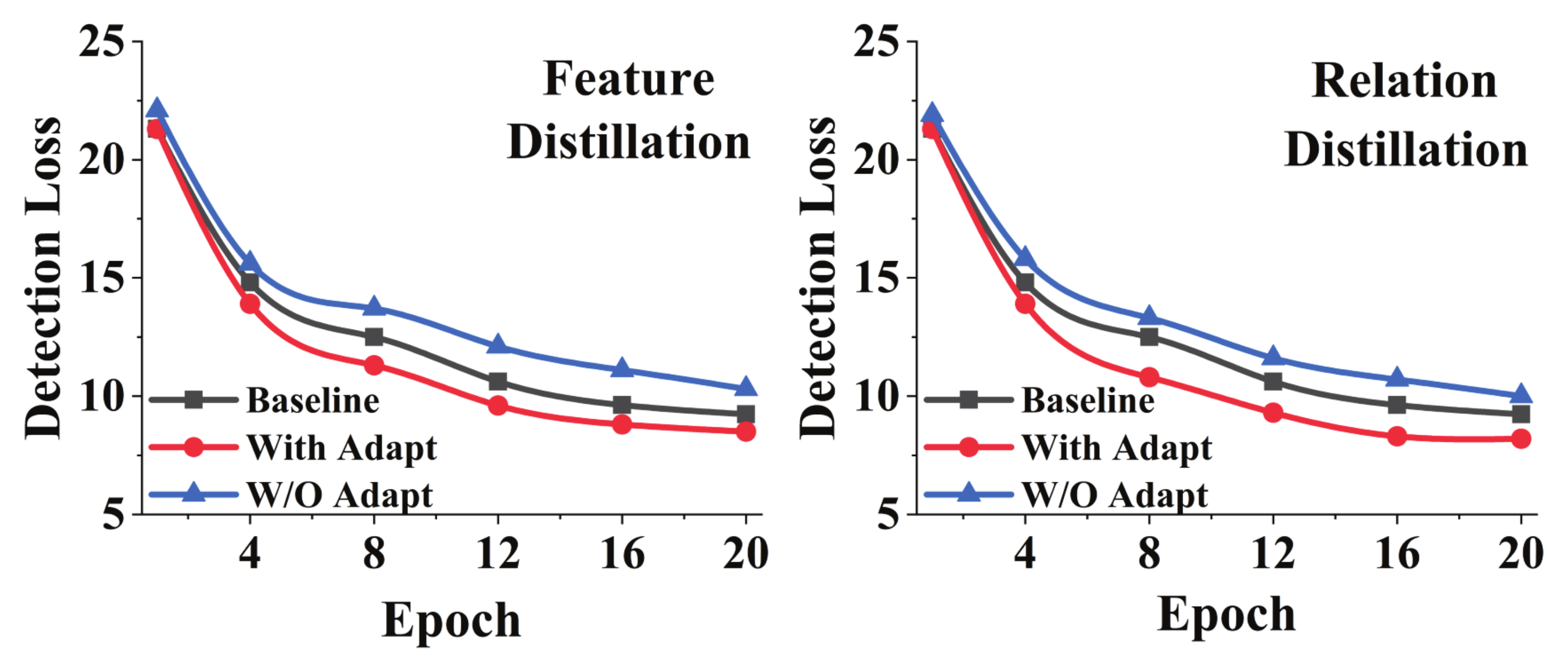}
\caption{Illustration to show that the adaptive layers help feature distillation and relation distillation decrease the detection loss.}
\label{img:4}
\end{figure}

We further compare the detection loss $\mathcal{L}_{\text{Det}}$ with/without the adaptive layers and the baseline is the student without UniDistill. The results in Figure \ref{img:4} show that, with the adaptive layers, the detection loss gradually becomes lower than the baseline. However, without the adaptive layers, the detection loss is always larger than the baseline, leading to worse performance. We think the problem results from that the feature quality of camera based teacher is worse than the LiDAR based student. Without the adaptive layers, aligning the features of student with teacher can decrease their quality. However, with the adaptive layers, the student can decide whether to learn from the teacher so that the performance is improved.
\subsection{Visualization}
In this section, we visualize the 3D object detection results to qualitatively show the effectiveness of UniDistill. The results are illustrated in Figure \ref{img:3}, where the teacher and the student are LiDAR-camera based and LiDAR based. From the results, the student detector can localize objects better with the help of UniDistill. Moreover, due to the balance between objects of different sizes, there are fewer false positive predictions on small objects.
	\section{Conclusion}
In this work, we propose a universal cross-modality knowledge distillation framework~(UniDistill) to improve the performance of single-modality 3D object detectors in BEV. UniDistill projects the features of  both the teacher and student into a unified BEV domain and then calculates three distillation losses to align the features for knowledge transfer. Taking advantage of the similar detection paradigm in BEV, UniDistill supports LiDAR-to-camera, camera-to-LiDAR, fusion-to-LiDAR and fusion-to-camera distillation paths. Furthermore, the proposed three distillation losses sparsely align foreground features to filter the misaligned background information and balance between objects of different sizes. Extensive experiments demonstrate that UniDistill is effective to improve the performance of student detectors. Inspired by block-wise distillation, we plan to leverage the distillation losses in a block-wise manner for acceleration, to further explore the potential of UniDistill.
        \clearpage
	{\small
		\bibliographystyle{ieee_fullname}    
		\bibliography{cvpr}

\begin{thebibliography}{10}\itemsep=-1pt

\bibitem{brazil2019m3d}
Garrick Brazil and Xiaoming Liu.
\newblock M3d-rpn: Monocular 3d region proposal network for object detection.
\newblock In {\em Proceedings of the IEEE/CVF International Conference on
  Computer Vision}, pages 9287--9296, 2019.

\bibitem{carion2020end}
Nicolas Carion, Francisco Massa, Gabriel Synnaeve, Nicolas Usunier, Alexander
  Kirillov, and Sergey Zagoruyko.
\newblock End-to-end object detection with transformers.
\newblock In {\em European conference on computer vision}, pages 213--229,
  2020.

\bibitem{chen2017learning}
Guobin Chen, Wongun Choi, Xiang Yu, Tony Han, and Manmohan Chandraker.
\newblock Learning efficient object detection models with knowledge
  distillation.
\newblock {\em Advances in neural information processing systems}, 30, 2017.

\bibitem{chen2020every}
Qi Chen, Lin Sun, Ernest Cheung, and Alan~L Yuille.
\newblock Every view counts: Cross-view consistency in 3d object detection with
  hybrid-cylindrical-spherical voxelization.
\newblock {\em Advances in Neural Information Processing Systems},
  33:21224--21235, 2020.

\bibitem{chen2017multi}
Xiaozhi Chen, Huimin Ma, Ji Wan, Bo Li, and Tian Xia.
\newblock Multi-view 3d object detection network for autonomous driving.
\newblock In {\em Proceedings of the IEEE conference on Computer Vision and
  Pattern Recognition}, pages 1907--1915, 2017.

\bibitem{chong2021monodistill}
Zhiyu Chong, Xinzhu Ma, Hong Zhang, Yuxin Yue, Haojie Li, Zhihui Wang, and
  Wanli Ouyang.
\newblock Monodistill: Learning spatial features for monocular 3d object
  detection.
\newblock In {\em International Conference on Learning Representations}, 2021.

\bibitem{dai2021general}
Xing Dai, Zeren Jiang, Zhao Wu, Yiping Bao, Zhicheng Wang, Si Liu, and Erjin
  Zhou.
\newblock General instance distillation for object detection.
\newblock In {\em Proceedings of the IEEE/CVF Conference on Computer Vision and
  Pattern Recognition}, pages 7842--7851, 2021.

\bibitem{fazlali2022versatile}
Hamidreza Fazlali, Yixuan Xu, Yuan Ren, and Bingbing Liu.
\newblock A versatile multi-view framework for lidar-based 3d object detection
  with guidance from panoptic segmentation.
\newblock In {\em Proceedings of the IEEE/CVF Conference on Computer Vision and
  Pattern Recognition}, pages 17192--17201, 2022.

\bibitem{guo2021distilling}
Jianyuan Guo, Kai Han, Yunhe Wang, Han Wu, Xinghao Chen, Chunjing Xu, and Chang
  Xu.
\newblock Distilling object detectors via decoupled features.
\newblock In {\em Proceedings of the IEEE/CVF Conference on Computer Vision and
  Pattern Recognition}, pages 2154--2164, 2021.

\bibitem{hinton2015distilling}
Geoffrey Hinton, Oriol Vinyals, Jeff Dean, et~al.
\newblock Distilling the knowledge in a neural network.
\newblock {\em arXiv preprint arXiv:1503.02531}, 2(7), 2015.

\bibitem{hou2020inter}
Yuenan Hou, Zheng Ma, Chunxiao Liu, Tak-Wai Hui, and Chen~Change Loy.
\newblock Inter-region affinity distillation for road marking segmentation.
\newblock In {\em Proceedings of the IEEE/CVF Conference on Computer Vision and
  Pattern Recognition}, pages 12486--12495, 2020.

\bibitem{hu2022afdetv2}
Yihan Hu, Zhuangzhuang Ding, Runzhou Ge, Wenxin Shao, Li Huang, Kun Li, and
  Qiang Liu.
\newblock Afdetv2: Rethinking the necessity of the second stage for object
  detection from point clouds.
\newblock In {\em Proceedings of the AAAI Conference on Artificial
  Intelligence}, volume~36, pages 969--979, 2022.

\bibitem{huang2021bevdet}
Junjie Huang, Guan Huang, Zheng Zhu, and Dalong Du.
\newblock Bevdet: High-performance multi-camera 3d object detection in
  bird-eye-view.
\newblock {\em arXiv preprint arXiv:2112.11790}, 2021.

\bibitem{huang2020epnet}
Tengteng Huang, Zhe Liu, Xiwu Chen, and Xiang Bai.
\newblock Epnet: Enhancing point features with image semantics for 3d object
  detection.
\newblock In {\em European Conference on Computer Vision}, pages 35--52, 2020.

\bibitem{ju2022paint}
Bo Ju, Zhikang Zou, Xiaoqing Ye, Minyue Jiang, Xiao Tan, Errui Ding, and
  Jingdong Wang.
\newblock Paint and distill: Boosting 3d object detection with semantic passing
  network.
\newblock In {\em Proceedings of the 30th ACM International Conference on
  Multimedia}, pages 5639--5648, 2022.

\bibitem{ku2018joint}
Jason Ku, Melissa Mozifian, Jungwook Lee, Ali Harakeh, and Steven~L Waslander.
\newblock Joint 3d proposal generation and object detection from view
  aggregation.
\newblock In {\em 2018 IEEE/RSJ International Conference on Intelligent Robots
  and Systems (IROS)}, pages 1--8. IEEE, 2018.

\bibitem{lang2019pointpillars}
Alex~H Lang, Sourabh Vora, Holger Caesar, Lubing Zhou, Jiong Yang, and Oscar
  Beijbom.
\newblock Pointpillars: Fast encoders for object detection from point clouds.
\newblock In {\em Proceedings of the IEEE/CVF conference on computer vision and
  pattern recognition}, pages 12697--12705, 2019.

\bibitem{li20173d}
Bo Li.
\newblock 3d fully convolutional network for vehicle detection in point cloud.
\newblock In {\em 2017 IEEE/RSJ International Conference on Intelligent Robots
  and Systems (IROS)}, pages 1513--1518. IEEE, 2017.

\bibitem{li2022uvtr}
Yanwei Li, Yilun Chen, Xiaojuan Qi, Zeming Li, Jian Sun, and Jiaya Jia.
\newblock Unifying voxel-based representation with transformer for 3d object
  detection.
\newblock In {\em Advances in Neural Information Processing Systems}, 2022.

\bibitem{li2022bevdepth}
Yinhao Li, Zheng Ge, Guanyi Yu, Jinrong Yang, Zengran Wang, Yukang Shi,
  Jianjian Sun, and Zeming Li.
\newblock Bevdepth: Acquisition of reliable depth for multi-view 3d object
  detection.
\newblock {\em arXiv preprint arXiv:2206.10092}, 2022.

\bibitem{liu2019structured}
Yifan Liu, Ke Chen, Chris Liu, Zengchang Qin, Zhenbo Luo, and Jingdong Wang.
\newblock Structured knowledge distillation for semantic segmentation.
\newblock In {\em Proceedings of the IEEE/CVF Conference on Computer Vision and
  Pattern Recognition}, pages 2604--2613, 2019.

\bibitem{liu2022bevfusion}
Zhijian Liu, Haotian Tang, Alexander Amini, Xinyu Yang, Huizi Mao, Daniela Rus,
  and Song Han.
\newblock Bevfusion: Multi-task multi-sensor fusion with unified bird's-eye
  view representation.
\newblock {\em arXiv preprint arXiv:2205.13542}, 2022.

\bibitem{liu2020tanet}
Zhe Liu, Xin Zhao, Tengteng Huang, Ruolan Hu, Yu Zhou, and Xiang Bai.
\newblock Tanet: Robust 3d object detection from point clouds with triple
  attention.
\newblock In {\em Proceedings of the AAAI Conference on Artificial
  Intelligence}, volume~34, pages 11677--11684, 2020.

\bibitem{luo2021m3dssd}
Shujie Luo, Hang Dai, Ling Shao, and Yong Ding.
\newblock M3dssd: Monocular 3d single stage object detector.
\newblock In {\em Proceedings of the IEEE/CVF Conference on Computer Vision and
  Pattern Recognition}, pages 6145--6154, 2021.

\bibitem{mao2021pyramid}
Jiageng Mao, Minzhe Niu, Haoyue Bai, Xiaodan Liang, Hang Xu, and Chunjing Xu.
\newblock Pyramid r-cnn: Towards better performance and adaptability for 3d
  object detection.
\newblock In {\em Proceedings of the IEEE/CVF International Conference on
  Computer Vision}, pages 2723--2732, 2021.

\bibitem{mao2021voxel}
Jiageng Mao, Yujing Xue, Minzhe Niu, Haoyue Bai, Jiashi Feng, Xiaodan Liang,
  Hang Xu, and Chunjing Xu.
\newblock Voxel transformer for 3d object detection.
\newblock In {\em Proceedings of the IEEE/CVF International Conference on
  Computer Vision}, pages 3164--3173, 2021.

\bibitem{pan20213d}
Xuran Pan, Zhuofan Xia, Shiji Song, Li~Erran Li, and Gao Huang.
\newblock 3d object detection with pointformer.
\newblock In {\em Proceedings of the IEEE/CVF Conference on Computer Vision and
  Pattern Recognition}, pages 7463--7472, 2021.

\bibitem{park2021pseudo}
Dennis Park, Rares Ambrus, Vitor Guizilini, Jie Li, and Adrien Gaidon.
\newblock Is pseudo-lidar needed for monocular 3d object detection?
\newblock In {\em Proceedings of the IEEE/CVF International Conference on
  Computer Vision}, pages 3142--3152, 2021.

\bibitem{philion2020lift}
Jonah Philion and Sanja Fidler.
\newblock Lift, splat, shoot: Encoding images from arbitrary camera rigs by
  implicitly unprojecting to 3d.
\newblock In {\em European Conference on Computer Vision}, pages 194--210,
  2020.

\bibitem{qi2018frustum}
Charles~R Qi, Wei Liu, Chenxia Wu, Hao Su, and Leonidas~J Guibas.
\newblock Frustum pointnets for 3d object detection from rgb-d data.
\newblock In {\em Proceedings of the IEEE conference on computer vision and
  pattern recognition}, pages 918--927, 2018.

\bibitem{romero2014fitnets}
Adriana Romero, Nicolas Ballas, Samira~Ebrahimi Kahou, Antoine Chassang, Carlo
  Gatta, and Yoshua Bengio.
\newblock Fitnets: Hints for thin deep nets.
\newblock {\em arXiv preprint arXiv:1412.6550}, 2014.

\bibitem{shi2020pv}
Shaoshuai Shi, Chaoxu Guo, Li Jiang, Zhe Wang, Jianping Shi, Xiaogang Wang, and
  Hongsheng Li.
\newblock Pv-rcnn: Point-voxel feature set abstraction for 3d object detection.
\newblock In {\em Proceedings of the IEEE/CVF Conference on Computer Vision and
  Pattern Recognition}, pages 10529--10538, 2020.

\bibitem{shi2019pointrcnn}
Shaoshuai Shi, Xiaogang Wang, and Hongsheng Li.
\newblock Pointrcnn: 3d object proposal generation and detection from point
  cloud.
\newblock In {\em Proceedings of the IEEE/CVF conference on computer vision and
  pattern recognition}, pages 770--779, 2019.

\bibitem{shi2020points}
Shaoshuai Shi, Zhe Wang, Jianping Shi, Xiaogang Wang, and Hongsheng Li.
\newblock From points to parts: 3d object detection from point cloud with
  part-aware and part-aggregation network.
\newblock {\em IEEE transactions on pattern analysis and machine intelligence},
  43(8):2647--2664, 2020.

\bibitem{tian2019fcos}
Zhi Tian, Chunhua Shen, Hao Chen, and Tong He.
\newblock Fcos: Fully convolutional one-stage object detection.
\newblock In {\em Proceedings of the IEEE/CVF international conference on
  computer vision}, pages 9627--9636, 2019.

\bibitem{vora2020pointpainting}
Sourabh Vora, Alex~H Lang, Bassam Helou, and Oscar Beijbom.
\newblock Pointpainting: Sequential fusion for 3d object detection.
\newblock In {\em Proceedings of the IEEE/CVF conference on computer vision and
  pattern recognition}, pages 4604--4612, 2020.

\bibitem{wang2021pointaugmenting}
Chunwei Wang, Chao Ma, Ming Zhu, and Xiaokang Yang.
\newblock Pointaugmenting: Cross-modal augmentation for 3d object detection.
\newblock In {\em Proceedings of the IEEE/CVF Conference on Computer Vision and
  Pattern Recognition}, pages 11794--11803, 2021.

\bibitem{wang2021fcos3d}
Tai Wang, Xinge Zhu, Jiangmiao Pang, and Dahua Lin.
\newblock Fcos3d: Fully convolutional one-stage monocular 3d object detection.
\newblock In {\em Proceedings of the IEEE/CVF International Conference on
  Computer Vision}, pages 913--922, 2021.

\bibitem{wang2019pseudo}
Yan Wang, Wei-Lun Chao, Divyansh Garg, Bharath Hariharan, Mark Campbell, and
  Kilian~Q Weinberger.
\newblock Pseudo-lidar from visual depth estimation: Bridging the gap in 3d
  object detection for autonomous driving.
\newblock In {\em Proceedings of the IEEE/CVF Conference on Computer Vision and
  Pattern Recognition}, pages 8445--8453, 2019.

\bibitem{wang2022detr3d}
Yue Wang, Vitor~Campagnolo Guizilini, Tianyuan Zhang, Yilun Wang, Hang Zhao,
  and Justin Solomon.
\newblock Detr3d: 3d object detection from multi-view images via 3d-to-2d
  queries.
\newblock In {\em Conference on Robot Learning}, pages 180--191. PMLR, 2022.

\bibitem{yan2018second}
Yan Yan, Yuxing Mao, and Bo Li.
\newblock Second: Sparsely embedded convolutional detection.
\newblock {\em Sensors}, 18(10):3337, 2018.

\bibitem{yang20203dssd}
Zetong Yang, Yanan Sun, Shu Liu, and Jiaya Jia.
\newblock 3dssd: Point-based 3d single stage object detector.
\newblock In {\em Proceedings of the IEEE/CVF conference on computer vision and
  pattern recognition}, pages 11040--11048, 2020.

\bibitem{yin2021center}
Tianwei Yin, Xingyi Zhou, and Philipp Krahenbuhl.
\newblock Center-based 3d object detection and tracking.
\newblock In {\em Proceedings of the IEEE/CVF conference on computer vision and
  pattern recognition}, pages 11784--11793, 2021.

\bibitem{yoo20203d}
Jin~Hyeok Yoo, Yecheol Kim, Jisong Kim, and Jun~Won Choi.
\newblock 3d-cvf: Generating joint camera and lidar features using cross-view
  spatial feature fusion for 3d object detection.
\newblock In {\em European Conference on Computer Vision}, pages 720--736,
  2020.

\bibitem{zhang2020improve}
Linfeng Zhang and Kaisheng Ma.
\newblock Improve object detection with feature-based knowledge distillation:
  Towards accurate and efficient detectors.
\newblock In {\em International Conference on Learning Representations}, 2020.

\bibitem{zheng2022boosting}
Wu Zheng, Mingxuan Hong, Li Jiang, and Chi-Wing Fu.
\newblock Boosting 3d object detection by simulating multimodality on point
  clouds.
\newblock In {\em Proceedings of the IEEE/CVF Conference on Computer Vision and
  Pattern Recognition}, pages 13638--13647, 2022.

\bibitem{zhou2018voxelnet}
Yin Zhou and Oncel Tuzel.
\newblock Voxelnet: End-to-end learning for point cloud based 3d object
  detection.
\newblock In {\em Proceedings of the IEEE conference on computer vision and
  pattern recognition}, pages 4490--4499, 2018.

\end{thebibliography}
	}
        \clearpage
        \twocolumn[
\begin{@twocolumnfalse}
\begin{center}
\Large{\textbf{Supplemental Material}}
\end{center}
\end{@twocolumnfalse}
]
This supplementary material contains additional details of the main manuscript. Section \ref{sec:1} presents additional details of the models and training strategies. Section \ref{sec:2} complements more experiments not included in the main manuscript. Section \ref{sec:3} shows more visualization results to prove the effectiveness of UniDistill.
\section{Details of Models and Training}
\label{sec:1}
To prove the effectiveness of UniDistill, we introduce BEVDet, CenterPoint and BEVFusion as the camera based, LiDAR based and LiDAR-camera based detectors. For BEVDet, the features of input images are firstly extracted by the backbone of ResNet-50 and then projected to BEV through LSS. We set the projected features to be the low-level BEV features $\boldmath{F}_{\text{cam}}^{\text{low}}$. With respect to CenterPoint, the input LiDAR points are distributed to regular voxels and the features of each voxel are extracted by 3D convolution. Then, the features of voxels in the same column are concatenated and we set the result as $\boldmath{F}_{\text{ldr}}^{\text{low}}$. BEVFusion builds on the above detectors  by concatenating $\boldmath{F}_{\text{cam}}^{\text{low}}$ with $\boldmath{F}_{\text{ldr}}^{\text{low}}$ and then processing it with a fully convolutional network~(FCN). The output features are set to be $\boldmath{F}_{\text{fuse}}^{\text{low}}$. The following steps are the same for different detectors, where a FCN follows as an encoder to produce $\boldmath{F}^{\text{high}}$ and then a detection head of CenterPoint generates classification and regression heatmaps. These heatmaps are used to form $\boldmath{F}^{\text{resp}}$.

For all detectors, during training, the detection head will calculate a classification loss $\mathcal{L}_{\text{Cls}}$ and a regression loss $\mathcal{L}_{\text{Reg}}$ that are combined to form the detection loss $\mathcal{L}_{\text{Det}}$. In Section 4.2 of the main manuscript, to help the detectors perform better, we use auto-scaling to balance the scales between $\mathcal{L}_{\text{Cls}}$ and $\mathcal{L}_{\text{Reg}}$ but turn it off in Section 4.3 for efficiency.

The training of detectors is finished on 20 GeForce RTX 2080Ti GPUs. These GPUs are distributed on 5 machines, where each machine has 4 GPUs, so that we adopt distributed training. Because of the limited memory, each GPU is distributed with 1 training sample.
\section{Complementary Experiments}
\label{sec:2}
In this section, experiments not included in the main manuscript are complemented. In Section \ref{sec:2.1}, UniDistill is compared with BEVDepth and MonoDistill to show its advantages. In Section \ref{sec:2.2.1}, the performance of UniDistill on Waymo is evaluated to show its generalization to different datasets. In Section \ref{sec:2.2}, we replace the detection head with a TransFusion based one and the backbone of BEVDet to Swin Transformer to show the generalization to different architectures. In Section \ref{sec:2.3}, more ablation studies about the adaptive layers and feature distillation are supplemented. In Section \ref{sec:2.4}, the training time and memory usage of UniDistill are listed.
\subsection{Comparison with BEVDepth and MonoDistill}
\label{sec:2.1}
To transfer the depth knowledge of LiDAR points to the camera based detector, which is BEVDet in our experiments, UniDistill introduces knowledge distillation for help. BEVDepth provides another approach for knowledge transfer by supervising the depth prediction of LSS in BEVDet with ground truth generated by projecting LiDAR points to the perspective view. Therefore, we compare UniDistill with BEVDepth to show the advantages of knowledge distillation. We build BEVDepth based on BEVDet by combining the detection loss $\mathcal{L}_{\text{Det}}$ with another depth prediction loss $\mathcal{L}_{\text{Depth}}$ and train it with the full training dataset. The performance of BEVDepth on the testing dataset is in Table \ref{supp_tab:1}. From the results, UniDistill helps BEVDet obtain better performance than BEVDepth, showing the advantages of knowledge distillation.
\begin{table}[t]\small
\centering
\caption{Comparison between UniDistill and BEVDepth on testing dataset to show the advantages of knowledge distillation. “L” and “C” represent LiDAR and camera.}
\resizebox{0.9\width}{!}{
\begin{tabular}{c|c|cclc}
\shline
Method     & \begin{tabular}[c]{@{}c@{}}Teacher\\ Modality\end{tabular} & mAP~$\uparrow$  & mASE~$\downarrow$ & mAOE~$\downarrow$ & NDS~$\uparrow$  \\ \shline
Baseline   & -                                                          & 26.4 & 26.6 & 55.8 & 36.1 \\
BEVDepth   & -                                                          & 28.4 & 26.3 & 55.3 & 37.7 \\ \shline
\rowcolor{black!5}UniDistill & L                                                          & \textbf{28.9} & \textbf{25.9} & \textbf{51.4} & \textbf{38.4} \\
\rowcolor{black!10}UniDistill & L+C                                                        & \textbf{29.6} & \textbf{25.7} & \textbf{49.2} & \textbf{39.3} \\ \shline
\end{tabular}
}
\label{supp_tab:1}
\end{table}
\begin{table}[t]\small
\centering
\caption{Comparison between UniDistill and MonoDistill on nuScenes test dataset. “L” and “C” represent LiDAR and camera.}
\resizebox{0.89\width}{!}{
\begin{tabular}{c|cc|ccc}
\shline
Method      & Modality           & \begin{tabular}[c]{@{}c@{}}Teacher\\ Modality\end{tabular} & mAP~$\uparrow$           & mASE~$\downarrow$          & NDS~$\uparrow$           \\ \shline
MonoDistill & \multirow{2}{*}{C} & \multirow{2}{*}{L}                                         & 23.2          & 28.7          & 34.3          \\ \cline{1-1} \cline{4-6} 
UniDistill  &                    &                                                            & \textbf{28.9} & \textbf{25.9} & \textbf{38.4} \\ \shline
\end{tabular}
}
\label{supp_tab:1.1}
\end{table}

\begin{table}[t]\small
\centering
\caption{Analysis to show the generalization of UniDistill to Waymo dataset. “L” and “C” represent LiDAR and camera.}
\resizebox{0.89\width}{!}{
\begin{tabular}{c|cc|ccc}
\shline
Method                      & Modality & \begin{tabular}[c]{@{}c@{}}Teacher\\ Modality\end{tabular} & mAPL~$\uparrow$           & mAPH~$\uparrow$          & \multicolumn{1}{l}{mAP~$\uparrow$} \\ \shline
\multirow{3}{*}{UniDistill} & L+C      & -                                                          &71.0               & 71.4              &75.3                          \\
                            & C        & -                                                          &22.3          & 33.0          & 34.5                     \\ \cline{2-6} 
                            & C        & L+C                                                        & \textbf{24.5} & \textbf{36.2} & \textbf{37.7}            \\ \shline
\end{tabular}
}
\label{supp_tab:2.1}
\end{table}

MonoDistill is another knowledge distillation framework that transfers the knowledge from a LiDAR-based teacher to a camera-based student. It directly unifies the architecture of the teacher and student by training the teacher with LiDAR points projected to the perspective view. Therefore, we further compare UniDistill with MonoDistill and the results are listed in Table \ref{supp_tab:1.1}, showing the better performance of UniDistill for the modality combination (C, L).

\subsection{Generalization to Waymo}
\label{sec:2.2.1}
In the main manuscript, all experiments are conducted on the nuScenes dataset. To show the generalization of UniDistill to different datasets, in this section, we further evaluate its performance on Waymo dataset. Specifically, UniDistill is first trained on the Waymo-mini dataset for 18 epochs and then tested on the whole validation set. The results in distillation path (2) are listed in Table \ref{supp_tab:2.1}, showing the effectiveness and generalization of UniDistill on Waymo.

\subsection{Generalization to More Architectures}
\label{sec:2.2}
In the main manuscript, we set the detection head of all detectors to be the same as that of CenterPoint. Therefore, we substitute it with a TransFusion based one and re-evaluate UniDistill to show the generalization of UniDistill to other  detection heads. The evaluation is conducted in distillation path (1) on the validation dataset and the modified detectors are trained on 1/2 training dataset for efficiency. Since the response distillation in UniDistill is not applicable to the TransFusion head, we only leverage the feature distillation and relation distillation for knowledge transfer. The results in Table \ref{supp_tab:2} reveal that UniDistill also improves the performance of the student detector, showing its generalization to different detection heads.

In addition, we substitute the ResNet-50 in BEVDet with Swin Transformer to show the generalization of UniDistill to different backbones. For efficiency, the modified detectors are trained on 1/2 training dataset and then evaluated  in distillation path (2) on the validation dataset. The results in Table \ref{supp_tab:2.2} show that UniDistill improves the performance of student detector and generalizes to different backbones.
\begin{table}[t]\small
\centering
\caption{Performance analysis to show the generalization of UniDistill to TransFusion. “L” and “C” represent LiDAR and camera.}
\resizebox{0.88\width}{!}{
\begin{tabular}{c|cc|ccc}
\shline
Method                       & Modality & \begin{tabular}[c]{@{}c@{}}Teacher\\ Modality\end{tabular} & mAP~$\uparrow$  & mASE~$\downarrow$ & NDS~$\uparrow$  \\ \shline
\multirow{2}{*}{TransFusion} & L+C      & -                                                          & 63.4 & 25.2 & 67.6 \\
                             & L        & -                                                          & 58.5 & 27.2 & 63.4 \\ \shline
UniDistill                   & L        & L+C                                                        & \textbf{60.9} & \textbf{25.9} & \textbf{65.9} \\ \shline
\end{tabular}
}
\label{supp_tab:2}
\end{table}
\begin{table}[t]\small
\centering
\caption{Analysis to show the generalization of UniDistill to Swin Transformer. “L” and “C” represent LiDAR and camera.}
\resizebox{0.92\width}{!}{
\begin{tabular}{c|cc|ccc}
\shline
Method                      & Modality & \begin{tabular}[c]{@{}c@{}}Teacher\\ Modality\end{tabular} & mAP~$\uparrow$           & mASE~$\downarrow$          & NDS~$\uparrow$           \\ \shline
\multirow{3}{*}{UniDistill} & L+C      & -                                                          & 63.3          & 24.7          & 69.0            \\
                            & C        & -                                                          & 27.8          & 27.6          & 36.0          \\ \cline{2-6} 
                            & C        & L+C                                                        & \textbf{32.5} & \textbf{25.7} & \textbf{39.7} \\ \shline
\end{tabular}
}
\label{supp_tab:2.2}
\end{table}

\subsection{Additional Ablation Studies}
\label{sec:2.3}
In Section 4.3.4 of the main manuscript, we conduct experiments to show that when evaluating in distillation path (3), the adaptive layers can avoid the performance degradation of the student after knowledge distillation. Some experiments in distillation path (4) are further designed to show that when the teacher detector performs better than the student, adopting the adaptive layers will decrease the effectiveness of UniDistill. The results are listed in Table \ref{supp_tab:3} and reveal that with the adaptive layers, the performance of the student slightly decreases. Therefore, when the teacher performs better than the student, there is no need to introduce the adaptive layers.

\begin{figure}[t]
\centering
\includegraphics[width=\linewidth]{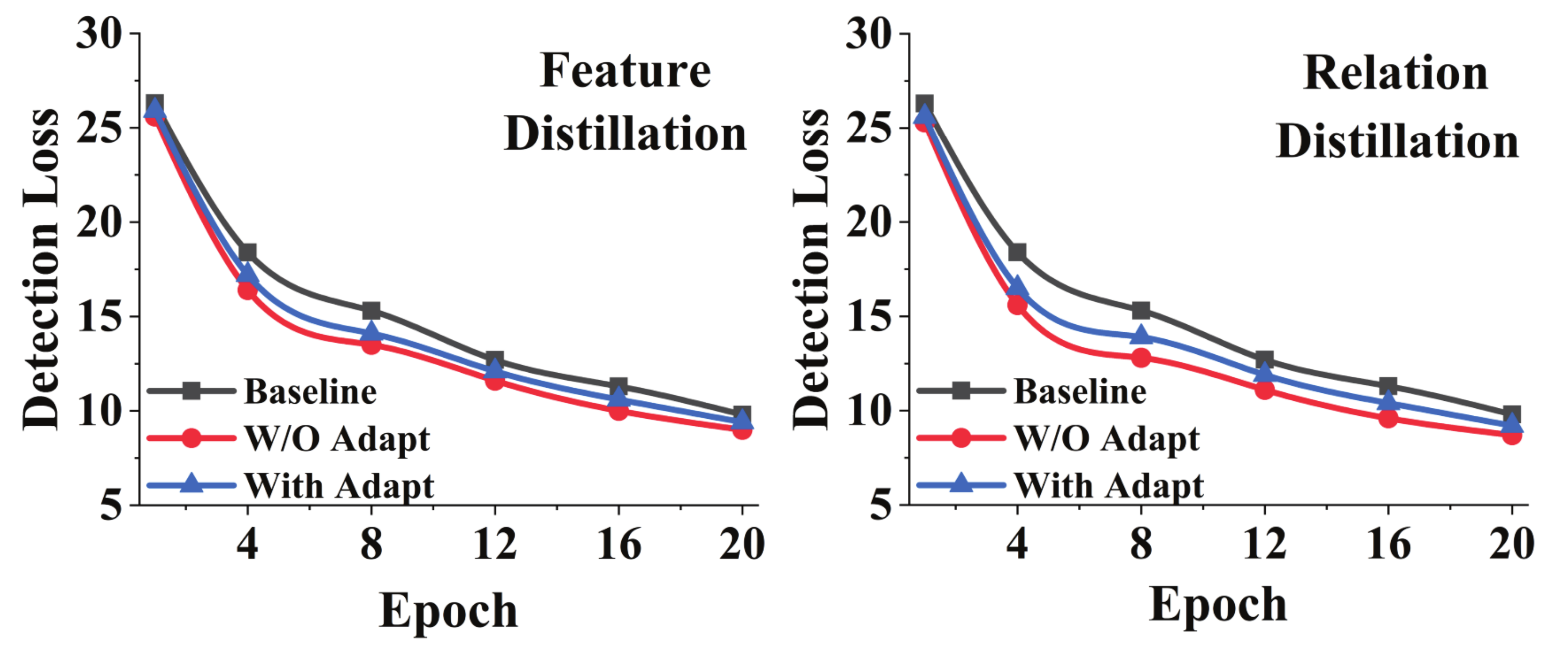}
\caption{Illustration to show that the adaptive layers increase the detection loss when the teacher performs better than the student.}
\label{supp_img:1}
\end{figure}

We also compare the detection loss $\mathcal{L}_{\text{Det}}$ with/without the adaptive layers and the baseline is the student without UniDistill. The results in Figure \ref{supp_img:1} show that with the adaptive layers, although the detection loss is lower than the baseline, it is always higher than that without the adaptive layers. We think the problem results from that the adaptive layers make it too free for the student to choose what to learn from the teacher. However, since the teacher detector is strong enough to instruct the student, directly aligning the features of the student with teacher can help the student learn better.

In addition, since most of the ablation studies are conducted in path (4), we complement the ablation studies in path (1) to improve the reliability.  As in Section 4.3.1 of the main manuscript,  we compare the original feature distillation with two modified ones that align the low-level BEV features (1) completely or (2) inside a Gaussian-like mask. The results are listed in Table \ref{supp_tab:3.1} and we can get the same conclusion that feature distillation performs better when selecting 9 crucial points for alignment.

\begin{table}[t]\small
\centering
\caption{Ablation study in path (4) to show that the adaptive layers decrease the effectiveness of feature distillation and relation distillation when the teacher detector performs better than student.}
\resizebox{0.81\width}{!}{
\begin{tabular}{c|ccc|ccc}
\shline
\multirow{2}{*}{Method} & \multicolumn{3}{c|}{$\mathcal{L}_{\text{Fea}}$} & \multicolumn{3}{c}{$\mathcal{L}_{\text{Rel}}$} \\ \cline{2-7} 
                        & mAP~$\uparrow$    & mAVE~$\downarrow$  & NDS~$\uparrow$   & mAP~$\uparrow$   & mAVE~$\downarrow$  & NDS~$\uparrow$   \\ \shline
Baseline                & 20.3   & 95.2  & 33.1  & 20.3  & 95.2  & 33.1  \\
With Adapt                & 20.7   & 91.4  & 33.9  & 21.2  & 89.6  & 34.2  \\ \shline
W/O Adapt              & \textbf{21.1}   & \textbf{88.5}  & \textbf{34.3}  & \textbf{21.7}  & \textbf{84.5}  & \textbf{35.0}  \\ \shline
\end{tabular}
}
\label{supp_tab:3}
\end{table}

\subsection{Training Time and Memory Usage}
\label{sec:2.4}
In this section, the training time and memory usage of detectors with/without UniDistill are listed. The detectors are trained on 1 GeForce RTX 2080Ti GPU and the training batch size is 1. For the training time, we list the average time to calculate the training loss. With respect to memory usage, we report the max allocated memory during training. The results are illustrated in Table \ref{supp_tab:4} and show that UniDistill will increase the training time and memory usage a lot. Therefore, we plan to introduce the block-wise distillation and other techniques to accelerate the training of UniDistill and decrease its memory usage.

\section{More Visualization Results}
\label{sec:3}
\begin{table}[t]\small
\centering
\caption{Ablation study in path (1) to show that feature distillation performs better when selecting crucial points for alignment.}
\resizebox{0.97\width}{!}{
\begin{tabular}{c|ccccc|c}
\shline
\multirow{2}{*}{Method} & \multicolumn{5}{c|}{AP~$\uparrow$}            & \multirow{2}{*}{NDS~$\uparrow$} \\ \cline{2-6}
                        & car  & truck & ped  & motor & mean &                      \\ \shline   
Baseline                    & 82.8 & 52.0  & 76.4 & 54.2  & 53.5 & 63.9                 \\            
Complete                    & 82.4 & 52.1  & 77.4 & 56.8  & 54.3 & 64.2                 \\
Gaussian                & \textbf{84.7} & \textbf{54.3}  & 76.1 & 53.4  & 54.7 & 64.8                 \\ \shline
Crucial                 & 82.9 & 50.5  & \textbf{82.4} & \textbf{61.7}  & \textbf{56.1} & \textbf{65.5}                 \\ \shline
\end{tabular}
}
\label{supp_tab:3.1}
\end{table}
\begin{table}[t]\small
\centering
\caption{Training time and memory usage of the detectors. “L” and “C” represent LiDAR and camera respectively.}
\resizebox{0.9\width}{!}{
\begin{tabular}{cc|cc}
\shline
Modality & \begin{tabular}[c]{@{}c@{}}Teacher\\ Modality\end{tabular} & Training Time~(s) & Memory Usage~(GB) \\ \shline
L+C      & -                                                          & 0.27                                                    & 5.96                                                   \\ \shline
C        & -                                                          & 0.13                                                    & 4.60                                                   \\
C        & L                                                          & 0.33~(+153\%)                                                    & 5.07~(+0.47)                                                   \\
C        & L+C                                                        & 0.40~(+207\%)                                                    & 6.44~(+1.84)                                                   \\ \shline
L        & -                                                          & 0.22                                                    & 3.21                                                   \\
L        & C                                                          & 0.46~(+109\%)                                                    & 4.51~(+1.30)                                                   \\
L        & L+C                                                        & 0.53~(+140\%)                                                    & 5.63~(+2.42)                                                   \\ \shline
\end{tabular}
}
\label{supp_tab:4}
\end{table}
\begin{figure*}[t]
\centering
\includegraphics[width=0.8\linewidth,height=9.5cm]{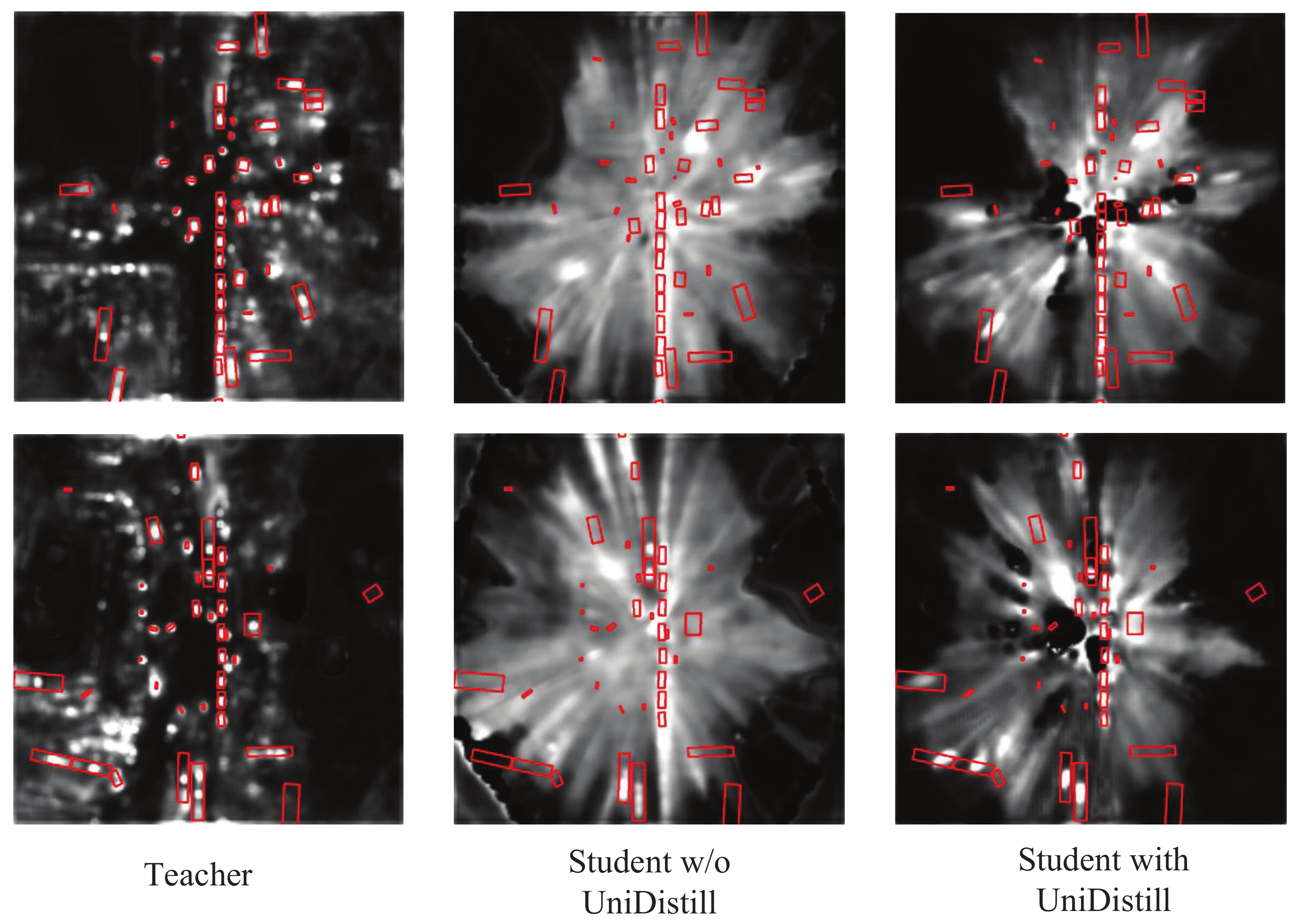}
\caption{Visualization of the response features. The boxes in red are the ground truth bounding boxes. The teacher and student detectors are LiDAR-camera based and camera based respectively. The first and second rows represent the results of two scenes. With UniDistill, the background areas are suppressed and the object boundaries are more clear.}
\label{supp_img:2}
\end{figure*} 
\begin{figure*}[t]
\centering
\includegraphics[width=0.8\linewidth,height=9.5cm]{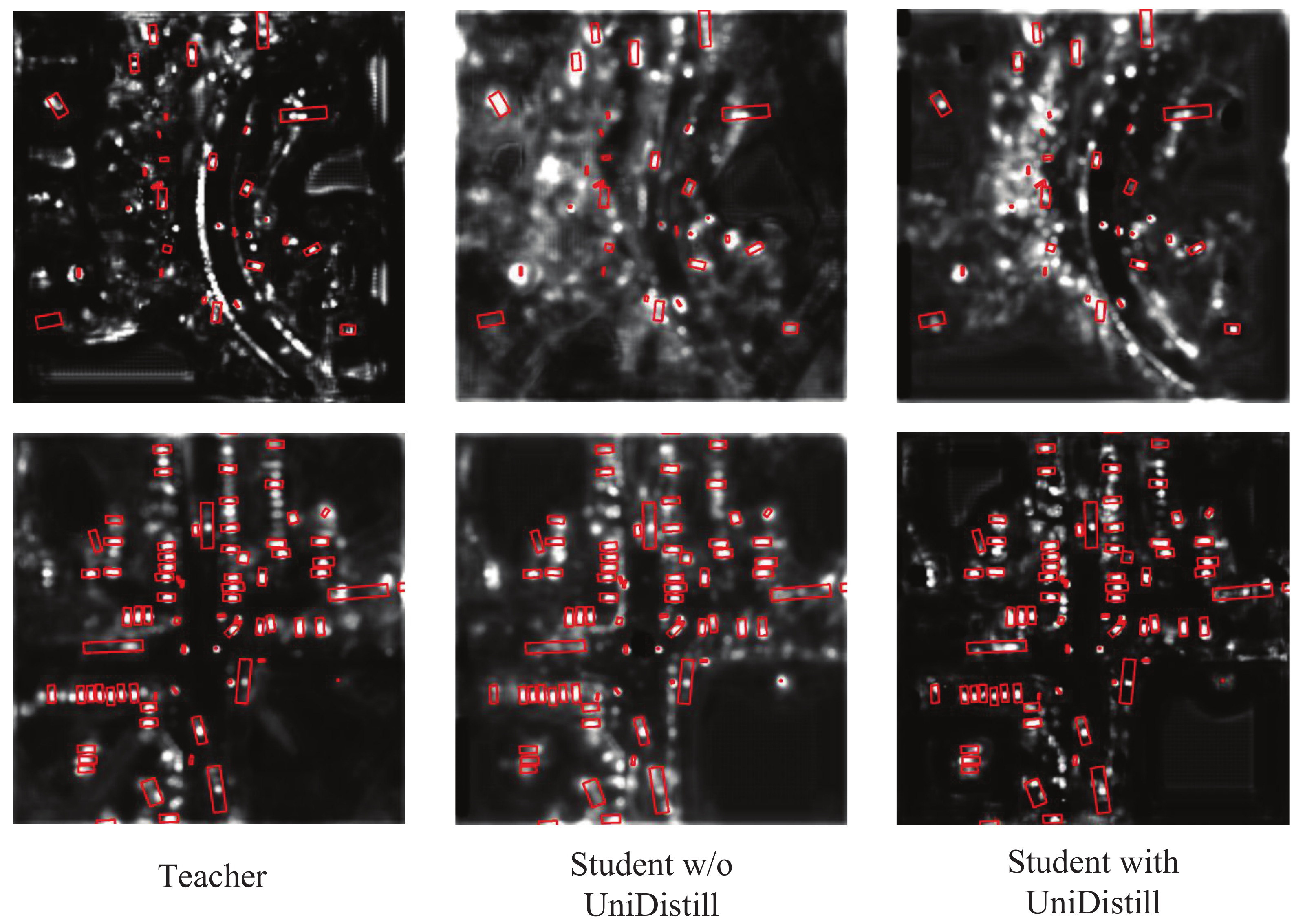}
\caption{Visualization of the response features. The boxes in red are the ground truth bounding boxes. The teacher and student detectors are LiDAR-camera based and LiDAR based respectively. The first and second rows show the results of two scenes. With UniDistill, the background areas are suppressed and the object boundaries are more clear.}
\label{supp_img:3}
\end{figure*} 
In this section, we provide more visualization results to show the effectiveness of UniDistill. For the response features of one teacher detector and one student with/without UniDistill, we calculate the mean along the channel dimension and visualize them. The results of the LiDAR-camera based teacher and the camera based student are illustrated in Figure \ref{supp_img:2} and that of the LiDAR-camera based teacher and the LiDAR based student are in Figure \ref{supp_img:3}. From the results, it is revealed that with UniDistill, the background areas are suppressed and the boundaries between objects are more clear. Therefore, there will be fewer false positive predictions and the detection performance is improved.

	\clearpage

\end{document}